\documentclass[3p,times]{elsarticle}

\usepackage{amssymb, amsmath}
\usepackage{graphicx}

\usepackage{epsfig}

\usepackage{multirow}
\usepackage{epstopdf}

\usepackage{url}
\usepackage{amsmath}
\usepackage{ecrc}
\usepackage[figuresright]{rotating}
\usepackage[T1]{fontenc}
\usepackage{amssymb}
\usepackage{amsmath}
\usepackage{amsfonts}
\usepackage{verbatim}
\usepackage{graphicx}
\usepackage{hyperref}
\usepackage{algorithmic}
\usepackage{algorithm}
\usepackage{array}

\usepackage{placeins}
\usepackage{xcolor,colortbl}

\volume{00}

\firstpage{1}

\journalname{Computer Vision and Image Understanding}

\jid{procs}

\jnltitlelogo{Computer Vision and Image Understanding}

\jnltitlelogo{Computer Vision and Image Understanding}

\begin{document}

\begin{frontmatter}

\dochead{}

%
\title{A Novel Approach for Human Action Recognition from Silhouette Images}

\author{Satyabrata Maity$^1$, Debotosh Bhattacharjee$^2$ and Amlan Chakrabarti$^1$
}

\address{1. A.K.Choudhury School of Information Technology,  
University of Calcutta, 92 A. P. C. Road, Kolkata:700 009, India.
\\2. Dept of Computer Science and Enginering,
Jadavpur University, 188, Raja S. C. Mallick Road, Kolkata, West Bengal 700032.
}


\begin{abstract}
 In this paper, a novel human action recognition technique from video is presented. Any action of human is a combination of several micro action sequences performed by one or more body parts of the human. The proposed approach uses spatio-temporal body parts movement (STBPM) features extracted from foreground silhouette of the human objects. The newly proposed STBPM feature estimates the movements of different body parts  for any given time segment to classify actions. We also proposed a rule based logic named rule action classifier (RAC), which uses a series of condition action rules based on prior knowledge and hence does not required training to classify any action. Since we don't require training to classify actions, the proposed approach is view independent. The experimental results on publicly available Wizeman and MuHVAi datasets are compared with that of the related research work in terms of accuracy in the human action detection, and proposed technique outperforms the others.

\end{abstract}


\end{frontmatter}

\section{Introduction}
The video based applications, which was once restricted for surveillance and entertainment purpose is now extended in every direction like education, health care,  social networking etc. This increasing demand directs a large number of researchers towards the domain of video analysis. Human action recognition (HAR) is the thrust to many applications like visual surveillance, video search and retrieval, human computer interaction and many more.

 Visual action recognition consists of different sub-topics such as gesture recognition developing human-computer interfaces, facial expression recognition  and abnormal activity detection for video surveillance. Conversely, full-body actions generally embrace a number of motions and require an integrated approach for recognition, encompassing facial actions, hand actions and feet actions.

Three main complexity issues, as mentioned in ~\cite{2}, are generally present in any HAR technique viz. i) Environmental complexity: The process of HAR depends on the quality of the video, which differs due to the environmental condition of the scene elements and makes the procedure more complex. This type of complexity includes occlusions, clutter, interaction among multiple objects, changing of illuminations etc. ii) Acquisition complexity: Besides environmental condition, the quality of the video also depends on video acquisition, which varies with respect to view point, movement of the camera etc. iii) Human action complexity: In general sense, human actions are of varied in nature, hence exact determination of human action is a complected task. Presence of multiple human entities makes any HAR technique more complex. So defining a model to recognize various kinds of human action is extremely difficult. To handle those three cases mentioned above, some constrains have been made in our proposed work viz. i) We have used the videos, which contain human silhouettes only and we didn't consider any silhouette extraction techniques also. video. So, the environmental and acquisition complicacies are not applicable for the proposed technique. ii) To reduce the complexity, the proposed work considers videos containing only one human object in each of the frame. iii) We also consider that the head should be in the upper portion of the body and the body should not be upside down. 
\begin{figure*}[!htb]
\centering
\includegraphics[scale=0.35]{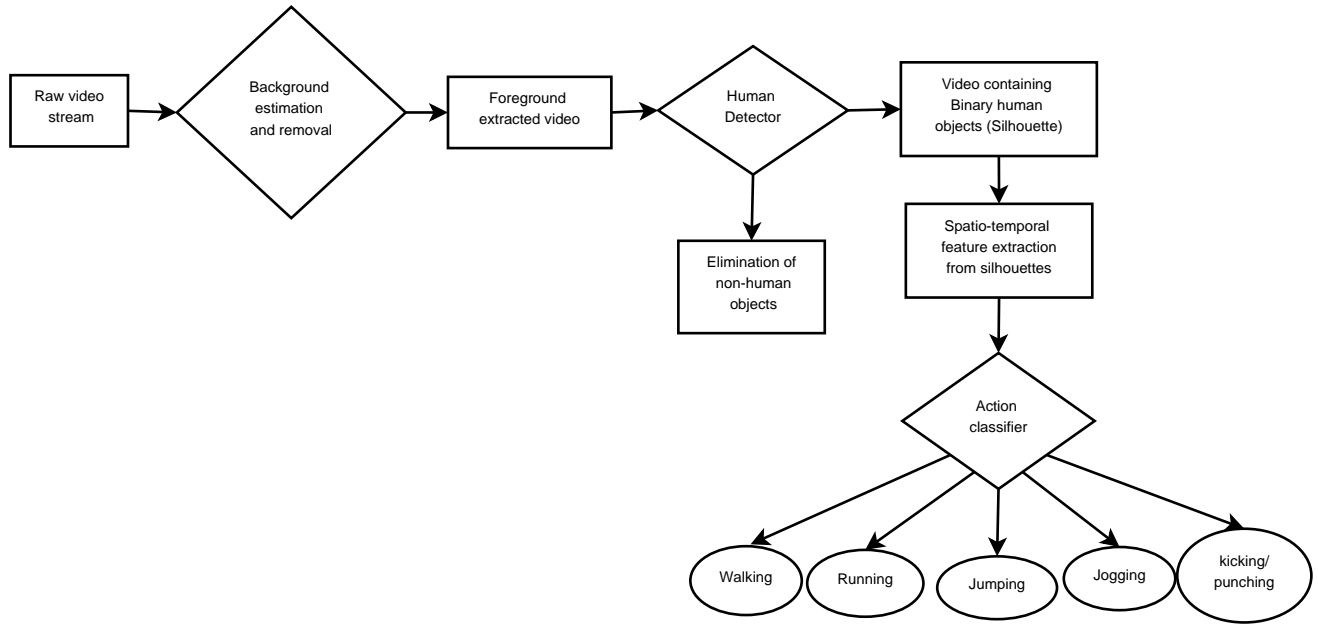}
\caption {Flow diagram of the general procedure}
\label  {fig:soa}
\end{figure*}

\par Fig.~\ref{fig:soa} illustrates the major components of a generic action recognition system using human silhouettes and their typical arrangement. 
The general frame-work for human action using silhouette analysis of consecutive frames of a video is mainly divided into three broad  sub-steps i) Foreground extraction: Foreground is extracted by eliminating background of the video. This step helps to reduce the searching area of the current frame. ii) Foreground classification: This step determines whether the foreground area contains human or not, as it is unnecessary to analyze nonhuman objects. iii) Feature extraction and action classification: Analysis of the movement of human body parts is performed for the consecutive frames to determine the human action in the successive frames. In our proposed work, we extracts spatio-temporal features from human silhouettes and then use the features for action classification. 

The actions performed by human are repetitive in nature and those actions are understandable by observing the movement of the body parts. The activity of these body parts determines action, so these body parts are termed as action units (AUs). In the proposed work, we have developed a rule based logic defined over different spatio-temporal values of AUs for determining numerous actions. 
The proposed approach selects four different parts of the body as AU viz. a) head, b) hand, c) leg and d) remaining body, since these four parts may be visible in a human silhouette. We design the algorithms to extract the location of these AUs in the frame containing human and then track them in the consecutive frames. We have introduced a new feature vector called Spatio-Temporal Body Parts Movement (STBPM), which contains location of a) HEAD, b) LEG(s), c) HAND(s) and d) Body Mass centre (BMC) and the derived features i.e. Head Angle (HA), Stride Angle (SA), Moving Direction (MD), Body Ratio (BR), and Bounding Box (BB). The dimension of the resulting feature vector is $k\times n$, where $k$ is the number of frames to recognize the action, so we can called it as action interval. $n$ is the number of features and the details of $k$ and $n$ will discuses in propose method section. Then Rule Action Classifier (RAC) is introduced to recognize the action. RAC is a classifier, which uses a series of condition action rules to categorize the action instead of any learning mechanism normally used by training based classifiers. 

 The main contributions of the proposed research work can be summarized as follows :
\begin{itemize}

\item Automatic localization of AUs viz. body, head, hand(s) and leg(s) automatically and the average accuracy rate is about $97\%$. 
\item Extraction of newly introduced low-dimensional STBPM features for human action recognition.
\item The proposed work is a first of the kind approach, which uses  a rule based logic set to determine different human actions.
\item The proposed method is view invariant except the top view, it can work in any viewpoint of the camera which makes our methodology more robust.
\end{itemize}

\par The paper is organized with seven sections starting with introduction section. Section-2 is related work, which contains a comprehensive study of the related research. The  description of the proposed methodology is included in section-3. Section-4 describes our results in details and analyses the efficiency of our results with compare to the results of other techniques of related research. Section-5, section-6 and section-7 include conclusion, acknowledgment and references respectively.
\section{Related work}

In the domain of computer vision, several communities are working to solve the difficulties of HAR in numerous ways. An outstanding survey of different approaches for HAR  are presented in ~\cite{1}. The HAR tasks can be roughly classified into five different categories which are based on shape models, motion models, geometric human body models, interest-point models, and dynamic models ~\cite{2}. 
The shape based models generally exploit the silhouette of moving objects commonly extracted by eliminating the backgrounds of the video frame. The silhouettes of the human object in consecutive frames are changing with respect to time as the movement is due to the displacement of one or more body parts. Some of the methods that apply shape based models are described in ~\cite{2}, ~\cite{3}, ~\cite{4}, ~\cite{5}, ~\cite{9}, ~\cite{14}, ~\cite{15}, ~\cite{16}, ~\cite{17}. Generally these methods use spatio-temporal features and in an ideal case, these approaches are invariant to luminance, color, and texture of the moving objects (and background) as mentioned in ~\cite{2}. All the approaches are heavily dependent on proper silhouette extraction of human objects, but accurate silhouette extraction in spite of color, texture and luminance is the main challenges in general. 

\par The second category described in ~\cite{18}, ~\cite{19}, ~\cite{20}, ~\cite{21} and ~\cite{22} generally apply flow based (mainly optical flow) techniques to model the motions of the moving objects in the consecutive frames of the video. A series of motion features of a moving object from the consecutive frames of a video provides the spatio-temporal model of any action. The flow-based method computes the optical field between adjacent frames and uses it as a feature for action recognition. This is suitable for recognizing small objects. However, it is computationally expensive and generate  coarse features only. Thus different actions may exhibit similar flows over short periods of time. 

The geometric human body models ~\cite{3}, ~\cite{23}, ~\cite{24}, ~\cite{25}, ~\cite{26}, attempt to characterize the geometric model of the human body to classify different actions as the geometry of the human body change according to the action he/she performs. This type of approaches construct feature vector, which contains static and dynamic geometric locations of different body parameters responsible to complete the action. Subsequently, a classifier is used to classify the extracted feature. This type of method can be very efficient if we correctly extract the geometric body parameters.


Point of interest models are described in ~\cite{27}, ~\cite{28}, ~\cite{29}, ~\cite{30} and ~\cite{31}. The points of interest are the distinguishable area of the image having eminent information. This type of approaches employ the points of interest collected from consecutive frames of the action interval and then process them to construct the feature vector. The classification is taken place exploiting this feature vector. ~\cite{29} and ~\cite{27} use 1D Gabor and Gaussian filter respectively to extract point of interest. On the other hand ~\cite{28} use Harris corner detector to extract salient points, which hold significant local variations in both spatial and temporal direction.  



The dynamic models ~\cite{32}, ~\cite{33}, classify human action based on the dynamic variation among the consecutive frames in temporal direction. This type of approaches characterize  the static posture of any action as state and then describing the dynamics using this state-space. Thus an action is a set of several states, the state space of which are connected using dynamic probability model (DPN). Hidden Markov Model (HMM) is the most commonly used DPN, which has the capability of directly modeling time variation of the data features.

\par Our proposed model combine the advantages of shape based model and geometric model to construct the feature vector. In our technique, we use silhouettes to extract spatio-temporal features, which contains the coordinates of different body-parts and some other derived features. Moreover, we designed a rule based logic to classify the human action, which is not required training.

\section{Proposed work}
\begin{figure}[!htb]
\centering
\includegraphics[scale=0.35]{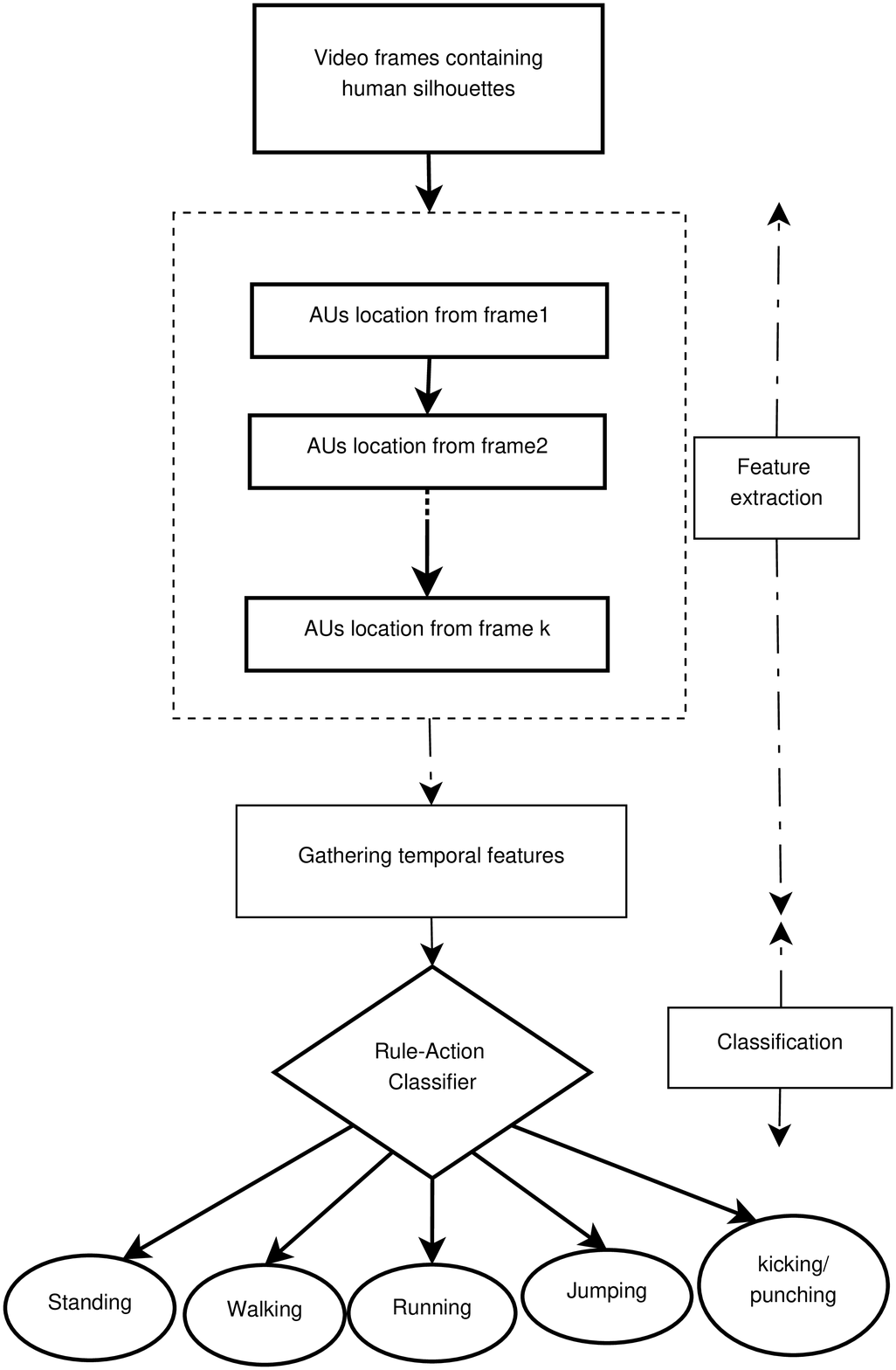}
\caption {Flow diagram of the proposed work}
\label  {fig:proposed_work}
\end{figure}

To be very precise the proposed work extracts STBPM features from human silhouettes of the consecutive frames of a video and then apply RAC to determine the action. Every shades of color add a dimension in the visual image information which is required to express an image more vividly but these shades of color do not have greater impact in HAR. On the other hand silhouette of a human object has only two levels and it provides us prominent shape based information. More specifically, silhouette of a particular frame of any human action presents the instance of the human shape of that time. Hence, the proposed approach exploits the silhouette towards action recognition, which is divided into two important steps A)Silhouette analysis and feature extraction, and B)Action classification. The block-diagram of the proposed work is shown in Fig.~\ref{fig:proposed_work}. First we analyze silhouettes of the human object in the first frame and compute all spatial feature values and store them in a row vector. The spatial features include the AUs location along with other derived features, which is discussed in the later of this section. The we process all the row vector of the action interval and compute the spatio-temporal features. The rule-action classifier to determine the action of the human in that action interval using spatio-temporal features.

\par
\begin{figure}[!htb]
\centering
\includegraphics[scale=0.35]{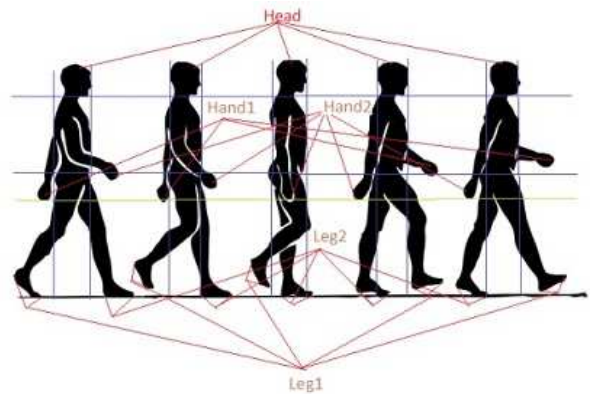}
\caption {Action unit of human body parts}
\label  {fig:AU}
\end{figure}

 Different AUs are involved in the various actions performed by human. The three most detectable AUs from human silhouettes, which are taking part of most of the actions, are head, hand(s) and leg(s). Fig.~\ref{fig:AU} shows an example which depicts these three AUs in case of walking. Every human body in the silhouette is divided with three horizontal and two vertical levels. The change in location or movement of AUs, in case of walking, can determine using these levels and the rules are proposed considering those movements. The following part of this section describes the process of feature extractions and action classification mechanism.

\subsection{Silhouette analysis and feature extraction} 
\begin{figure}[!htb]
\centering
\hspace{40pt}
\includegraphics[scale=0.45]{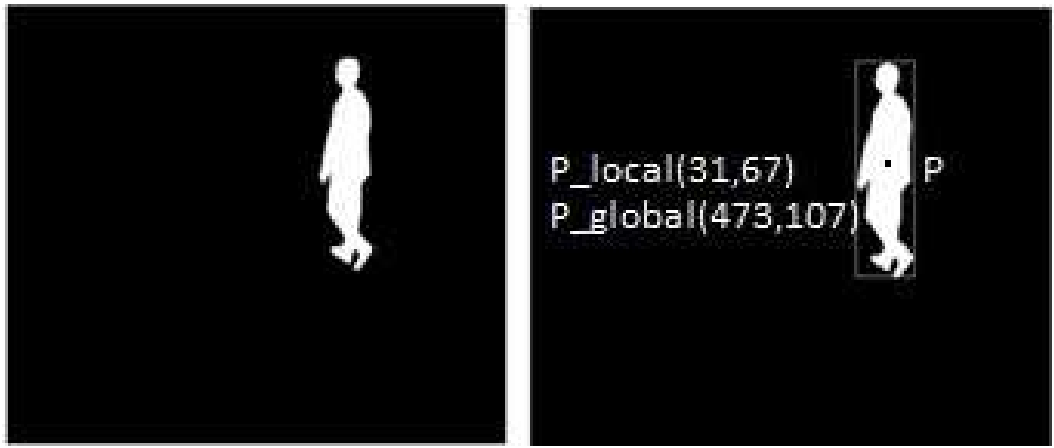}
\caption {Algorithm of BMC computation}
\label  {fig:LM_GM}
\end{figure}

\begin{figure}[!htb]
\centering
\includegraphics[scale=0.32]{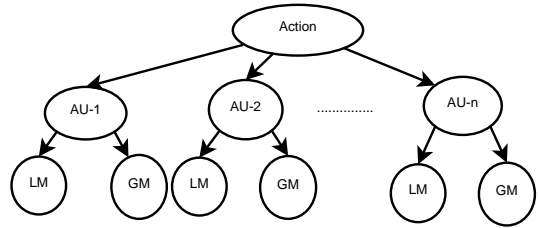}
\caption {Hierarchy of human an action}
\label  {fig: hierarchy}
\end{figure}

In the proposed work, we define actions in terms of AUs and their movement, which is needed to perform the action. The movement of AUs are either i) local movement (LM): movement with the BBOX, ii) global movement (GM): movement with in the frame. An idea of local and global position of AU are shown in Fig.~\ref{fig:LM_GM}, Where the point P has two coordinates i) P\_local to represent the local coordinate with respect to BBOX, and ii) P\_global to represent the global coordinate with respect to frame. This hierarchy of human action as shown in Fig.~\ref{fig: hierarchy} is characterized in rule base logic for HAR.   

The objective of this step is generation of STBPM feature vector, which holds different state of AUs. The feature vector contains location and other relevant information from the silhouettes of the $k$ consecutive frames of any video, which are considered for taking the decision of any human action. In our experiments, we have set the action interval $k$= number of frames/sec i.e. the frame rate of the video. The STBPM feature vector is formed with $28$ different values viz.
\begin{itemize}
\item Bounding box $(BBOX)$ with four values ($x_{st}$, $y_{st}$, $dx$ and $dy$).
\item Body mass center $(BMC)$ with two values ($x_{cnt}$, $y_{cnt}$).
\item Global body mass center $(BMC_G)$ with two values ($xg_{cnt}$,$yg_{cnt}$).
\item Body Ratio $(BR)$ with one value.

\item Extreme left and extreme right sided locations of head with four values (($HD1_x$, $HD1_y)$ \& ($HD2_x$, $HD2_y$)).
\item Angle of head $(HA)$ with one value measured in form of degree, which is discuss in the later part of this section.
\item $HAND1$ and $HAND2$ with four values (($xh1$,$yh1$) \& ($xh2$,$yh2$)).
\item $HEEL1$, $TOE1$, $HEEL2$ \& $TOE2$ with eight values ($xhl1$,$yhl1$), ($xt1$,$yt1$), ($xhl2$,$yhl2$) \& ($xt2$,$yt2$)). 
\item Moving direction ($MD$) with one value.
\end{itemize}

These $28$ values extracted each from foreground silhouettes of $k$ consecutive frames make the size of the feature vector equal to $k \times 28$ for each action interval. Thus two levels of features are extracted for classification 1)Spatial features: The Location of AUs and other derived feature extracted from each human silhouette. 2) Generation of spatio-temporal features: The collection of spatial features from action interval of consecutive frames and then reduce the dimensionality of the huge number of features. In the feature vector, all the locations are represented using 2-D spatial co-ordinate in the form of $(x,y)$. 
 
\subsubsection {Spatial features}
The spatial features include the spatial location of AUs and their derived features. So, we divided the process of spatial feature extraction into four phases viz.
\begin{itemize}
\item Extraction of BBOX, BMC, BMC\_G \& BR.
\item Extraction of head locations and HA.
\item Extraction of hand(s) location(s).
\item Extraction of leg(s) location(s) and SA.
\end{itemize}
 Before describing the details of extraction of spatial features in details, we explain four vectors viz. X1, X2, Y1 and Y2, which are used to extract the location of AUs and other derived features. Silhouette image has only two levels viz. $'0'$ and $'1'$. So there is no information in the inner region of the silhouette. Thus the analysis is done on the boundary region of the silhouette. The proposed methodology extract four extreme bounding vectors, two of them along the row viz. $X_1$ and $X_2$ and other two are along the column viz. $Y_1$, $Y_2$. We named the 2-D matrix containing the silhouette of a human object as SIL.

$X_1$ vector characterizes the extreme left-sided boundary curve of the human silhouette. The left-sided curve contains the first nonzero point of each row of the human silhouette as computed in Eq-1.
 
\begin{equation}
X_1(i)=j, IF\hspace{.2cm} SIL(i,j)=1, \& \hspace{.2cm}\forall \hspace{.08cm}SIL(i,l)=0.\\
\end{equation}
Where, $l \textless j$.
 
$X_2$ vector characterizes the extreme right-sided curve of the human silhouette. The right-sided curve contains the last nonzero point of each row of the human silhouette as shown in Eq-2.

\begin{equation}
X_2(i)=j, IF\hspace{.2cm} SIL(i,j)=1, \& \hspace{.2cm}\forall \hspace{.08cm}SIL(i,l)=0.\\
\end{equation}
Where, $l \textgreater j$.

$Y_1$ vector characterizes the peak curve of the human silhouette. The peak curve contains the starting nonzero point of each column of the human silhouette as computed in Eq-3.

\begin{equation}
Y_1(i)=j, IF\hspace{.2cm} SIL(j,i)=1, \& \hspace{.2cm}\forall \hspace{.08cm}SIL(l,i)=0.\\
\end{equation}
Where, $l \textless j$.

$Y_2$ vector characterizes the base curve of the human silhouette. The base curve contains the ending nonzero point of each column of the human silhouette as computed in Eq-4.

\begin{equation}
Y_2(i)=j, IF\hspace{.2cm} SIL(j,i)=1, \& \hspace{.2cm}\forall\hspace{.08cm} SIL(l,i)=0.\\
\end{equation}
Where, $l \textgreater j$.

\begin{itemize}
\item{Extraction of BBOX, BMC,BMC\_G \& BR features values:} 
$BBOX$ is represented by $(x_{st},y_{st}, dx, dy)$ where $(x_l,y_l)$ is the left top point and $dx$ \& $dy$ are the width and height of the BBOX and all the values are computed using four vectors as described in Eq.5 to Eq.12. Now we have two windows viz. BBOX and the image frame. The co-ordinate values with respect to BBOX are the local values and the same with respect to image frame are the global values. The conversion of any local co-ordinate to its corresponding global value can be easily done by adding $x_{st}$ and $y_{st}$ values with the corresponding x and y values of the co-ordinate and the opposite mechanism is done to convert the global co-ordinate to its local one. This conversion is needed to estimate global movements of any AU.
\begin{equation} 
x_{min}=MIN(X_1).
\end{equation}
\begin{equation}
y_{min}=MIN(Y_1).
\end{equation}
\begin{equation}
x_{max}=MAX(X_2).
\end{equation}
\begin{equation}
y_{max}=MAX(Y_2).
\end{equation}
\begin{equation}
x_{st}=x_{min}.
\end{equation}
\begin{equation}
y_{st}=y_{min}.
\end{equation}
\begin{equation}
dx= x_{max}- x_{min}.
\end{equation}
\begin{equation}
dy= y_{max}- y_{min}.
\end{equation}

Body mass center ($BMC$) is a key parameter for action detection. The location of $BMC$ in consecutive frames determines several local movements. BMC is computed using weight of rows and columns. The weight of each row or column is the total number of nonzero values along the respective rows or columns. ($x_{cnt}$ and $y_{cnt}$) are the weighted average of all rows  and all columns of the BBOX respectively. The proposed work use Eq-13 to Eq-16 to extract BMC. Fig.~\ref{fig:BMC} shows the BMC and the contour of corresponding human silhouette.
\begin{equation}
ROW_{wt}(i)=\sum_{j=1}^{col}SIL(i,j)
\end{equation}

\begin{equation}
COL_{wt}(i)=\sum_{j=1}^{row}SIL(j,i)
\end{equation}
 
\begin{equation}
x_{cnt}=(\sum_{i=1}^{row}ROW_{wt}(i)\times i)/\sum_{i=1}^{row}row_{wt}(i)
\end{equation}

\begin{equation}
y_{cnt}=(\sum_{i=1}^{col}COL_{wt}(i)\times i)/\sum_{i=1}^{col}COL_{wt}(i)
\end{equation}

The global body mass center(BMC\_G) is represented by $(xg_{cnt},yg_{cnt})$ is computed using Eq.13 and Eq.14. 
\begin{equation}
xg_{cnt}=x_{cnt}+x_{st}.
\end{equation}
\begin{equation}
yg_{cnt}=y_{cnt}+y_{st}.
\end{equation}
Most of the feature values are computed in the proposed methodology in their respective local windows i.e BBOX as we need to tract the movement of any  AU. The movement of human objects should reflect in the BMC values of consecutive frames, but it is not always reflected by its local coordinate values. In some cases, the values of BMC remain nearly unchanged inspite of huge movement in that action interval. For instance, the location value of BMC and HEAD change a little with respect to BBOX in case of "running". The value of BMC\_G has a big role in such cases as the values of BMC\_G of the corresponding frames are changed appreciably if there is any global movement. 
 
Human body ratio is the ratio of the height and the width of the human body as computed in Eq.15.
\begin{equation}
BR=dx/dy
\end{equation} 
The anatomical studies done in ~\cite{36} proposed the vertical position of head and hip are $0.87H$ and $0.53H$ respectively, where H is the height(from head to toe) of the human body. The ending point of hand can go little beyond hip if we simply keep our hand down and straight and hence, we consider this as $0.47H$. We divide the BBOX in smaller regions using three horizontal and two vertical lines. i) Upper level (UL): The line, which separates upper region of the body from in the BBOX. In Fig.~\ref{fig:leveling}, the horizontal solid line in the top area of the BBOX is the UL and the region above this line is upper region. Upper region contains head and hand(s) (when someone uplift his/her hand) and the line is at $0.87H$. ii) Middle level (ML): This is a line, which defines the middle region in the BBOX. The middle region, which contains body and hands, is started from UL and end at ML. ML is the dashed horizontal line at $0.47H$ as shown in Fig.~\ref{fig:leveling}. iii) Lower level (LL): Human legs are included in the region below LL and the line is at $0.53H$ as shown in Fig.~\ref{fig:leveling} lower level in dotted line. Some times hand(s) and leg(s) share a common portion in BBOX, a common area between lower and middle regions is there for that reason.  iv) Left side level (LSL): This is a vertical line with a distance of v1 from the BMC. v1 is the average of all values of X1 with in ML. v) Right sided level (RSL): This is computed in same manner with LSL except taking the values from X2. In Fig.~\ref{fig:leveling}, left sided vertical line is LSL and right sided vertical line is RSL.

\begin{figure}[!htb]
\centering
\hspace{20pt}
\includegraphics[scale=0.30]{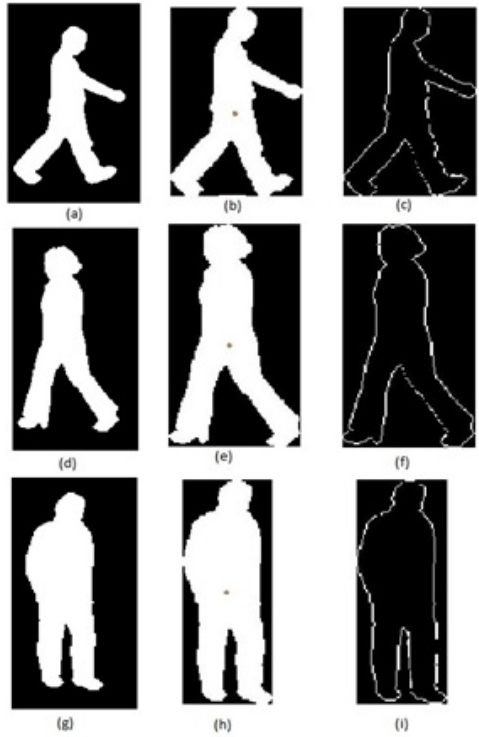}
\caption {Flow diagram of the proposed method}
\label  {fig:BMC}
\end{figure}

\begin{figure}[!htb]
\centering
\includegraphics[scale=0.30]{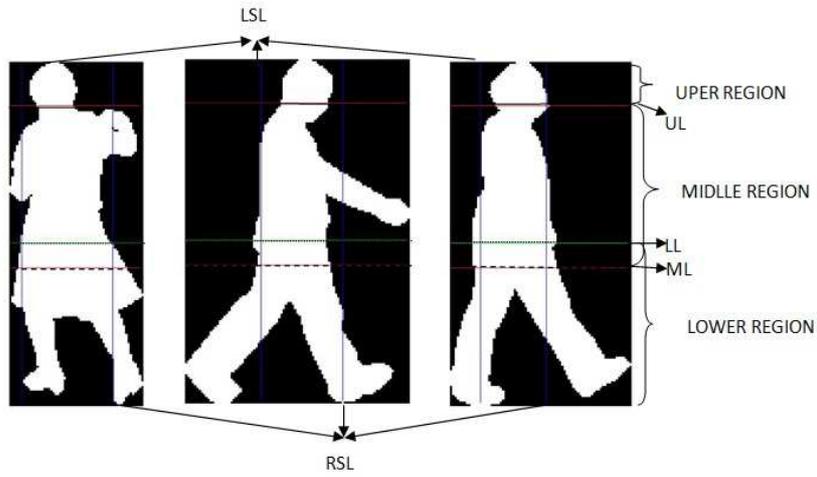}
\caption {Three logical horizontal level of human silhouette}
\label  {fig:leveling}
\end{figure}

\item {Extraction of head locations and HA:}  Head, located above the UL, is the highest part of the body unless one raises hand(s) in most of the cases of common human action like walking, running; jumping, jogging etc. On the other hand the vector $Y1$ contains the starting points from the top of BBOX. Head, hand or other body parts have certain shape. So the $Y1$ curve contains several numbers of curvatures, which differ depending upon different shape of the different body parts.

We have from Eq.20, $Y'_1$ contains the column having Y1 value greater than $UL$ mark. From have from Eq.21, the component array $COMP$ consists the length of the components present in the upper level. The highest length component is the HEAD and the most left sided point of the HEAD is starting point $H_{st}$ and the most right sided point of the HEAD ending point $H_{en}$.


\begin{equation}
\resizebox{.25\hsize}{!}{ $Y'_1(i)=\left\{\begin{array}{ll}\mbox{$1$} & \mbox{if $Y'_1(i) > UL$};\\
\mbox{$0$} & \mbox{$Otherwise$}.\\
\end{array}\right. $}
\end{equation}

\begin{equation}
COMP(i)=\sum_{j=k1}^{k2}Y'_1(j) \mbox{where $\forall Y'_1(j)>0$}\\ 
\mbox {\& $ k1\leq  j \leq k2$ }
\end{equation}

Fig.~\ref{fig:head_find} shows some results after applying HEAD extraction formula using Eq.20 and Eq.21. In the figure two end points $H_{st}$ and $H_{en}$ are connected with the BMC of the corresponding human silhouette.

\begin{figure}[!htb]
\centering
\includegraphics[scale=0.25]{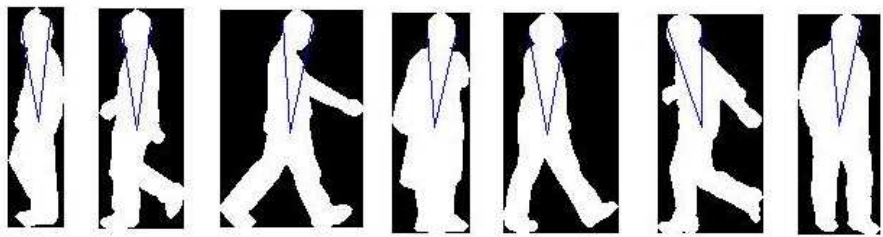}
\caption {Locating head from BMC}
\label  {fig:head_find}
\end{figure}

HA is the angle between head and the shoulder. HA is required to determine the body moving direction of the human as the head is angled towards the moving direction in most of the cases. Head angle (HA) is measured by using the middle point ($H_{mid}$) of the head and the middle point of the shoulder ($SH_{mid}$) . The location of shoulder is computed by standard anatomical height of $0.818\times H$, where H is the height of the human body (from lowest heel to top point of the head). 
\begin{equation}
\resizebox{.5\hsize}{!}{ $H_{mid}(x_1,y_1)=\left\{\begin{array}{ll}\mbox{$x_1=(X_1(H_{st})+X_2(H_{mid}))/2$} & \\
\mbox{$y_1=(H_{st}+H_{en})/2 )$}.\\
\end{array}\right. $}
\end{equation}

\begin{equation}
\resizebox{.5\hsize}{!}{ $Sh_{mid}(x_2,y_2)=\left\{\begin{array}{ll}\mbox{$x_2=(X_1(H\times 0.818)+X_2(H\times 0.818))/2$} & \\
\mbox{$y_2=(H\times 0.818 )$}.
\end{array}\right. $}
\end{equation}
\begin{equation}
HA=\tan^{-1}\left[(y_2-y_1)/(x_2-x_1) \right]
\end{equation}

\item{Hand localization:} Unlike head, hand(s) can be in two regions above UL and ML. Sometimes it is occluded by different body area. It can be tracked when it is appearing outside the body area. Two vectors X1 and X2, which hold the left and the right sided points of the silhouette must contain the information about the hand. 

\begin{equation}
std_{X1}=STD( X'_1(i))   \mbox{ $where X'_1=X_1(LL:UL)$}
\end{equation}

\begin{equation}
X"_1(i)=X_1(i) \mbox{ $\forall abs(X_1(i)-BMC(x))\geq std_{X1} $}
\end{equation}

\begin{figure}[!htb]
\centering
\includegraphics[scale=0.25]{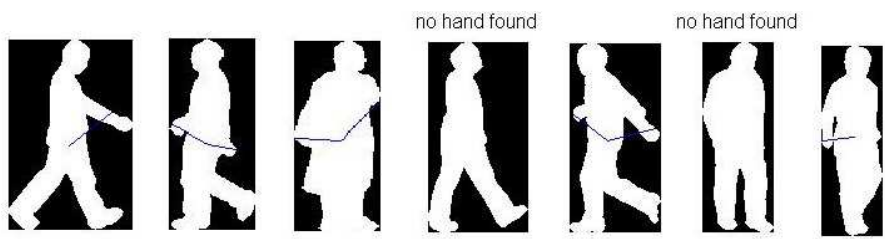}
\caption {Locating hand(s) from BMC.}
\label  {fig:hand_find}
\end{figure}

The area beyond two vertical lines contain the information of hand. Thus we analyze   those area stored in $x"_1$ of Eq. 21. If there is any area beyond hand determine threshold $Th_{hnd}$ and sustain up to $0.03H$, then there is a hand. This approach is applied on $x_2$ to get second hand if visible.  

\item{Heel and toe localization and stride angle computation:} This step includes most important feature of human action recognition as most of the active action are depending on the different movement of legs. The action area is performing below the LL line of the BBOX. $Y2$ vector is exploited to find the position of leg and corresponding stride angle (where two legs are detected). The proposed method extract the position of toe and heel of the leg(s) as it is important to understand the direction of human motion. In natural view the human motion is along the direction of the toe and the heel which holds the body weight remaining the back side. Eq-5 is used to extract the bottom point(s) of $Y2$ curve. $bs\_pnt$  is used  instead of $pk\_pnt$ and loc are two vectors contain bottom value(s) and corresponding location of the curvatures. One more function $leg\_find$ analyzes the shape of the curvature and return the toe and hill vectors.

\begin{equation}
Th_l=mean( Y'_2(i))   \mbox{where $Y'_2(i)) \in \{ x\|\forall Y_2(i)>LL\}$}
\end{equation}

\begin{equation}
Y"_2(i)=Y_2(i) \mbox{ $\forall abs(Y_2(i)-BMC(y))\geq Th_l $}
\end{equation}

If only one leg found, it is assumed that stride angle is zero as one leg is occluded by others; otherwise the angle between two legs is measured and stored in $SA$. Fig.~\ref{fig:leg_find} shows some results of leg finder where the heel(s) and toe(s) are connected with the BMC.

\begin{figure}[!htb]
\centering
\includegraphics[scale=0.25]{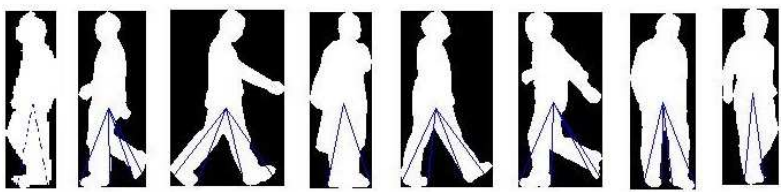}
\caption {Locating leg(s) from BMC.}
\label  {fig:leg_find}
\end{figure}


\end{itemize}

\subsubsection{Generation of spatio-temporal features} 
\par
\textit{Extraction of moving direction (MD):} MD is temporal feature, which is extracted  by exploiting the extracted values of BMC\_G. BMC\_G is of two types i) Static: there is no such change in BMC\_G, ii) Dynamic: The value of BMC\_G is changing with respec to time. In case of static BMC\_G, MD is determined using HA and heel-toe locations of the last frame. If the direction is close to $90^{\circ}$  degree refers to non-moving head; otherwise  i)Angle greater than $95^{\circ}$  refers the movement is towards right. or ii)Angle less than $85^{\circ}$  refers to left sided movement. The locations of heel(s) and toe(s) also have contribution to determine the MD. If TOE(s)>HEEL(s) with respect to Y values, then the direction is towards right and vice-versa. Undefined refers to the fact when leg is not correctly defined or HEEL(s)-TOE(s) are not correctly extracted i.e. smaller or bigger than normal or one HEEL is  greater than TOE and other TOE is greater than HEEL. In the cases HA and TOE-HEEL provide different values, the value of the other is taken if one is zero, else zero is taken.  On the other hand MD is determined by the average value of BMC\_G features. The direction is computed by comparing the value of average value of BMC\_G with the starting value of BMC\_G. The direction has any of the three values '-1' for left direction, '0' for straight and '1' for right direction.

The feature values as depicted in the spatial feature section are extracted from $k$ number of consecutive frames. Thus we have the feature size $k \times 28$. But all AUs may not active for every work. So we can eliminate the non-active features to take the classification decision in faster way. We have now $28$ columns and each one contains $k$ number of values. The average(AV), min (MIN), max (MAX) and standard deviation (STD) values from each column a are extracted. In case of nonactive feature, the values AV, MIN and MAX are same and STD values are near to zero. Now we have a resultant feature vector of size $k \times 28$ contains AV, MIN, MAX and STD of all feature values. This is used to classify the action. 

\subsection{Action Classification}
The successful completion of previous step, we got a feature vector, which contains the activity of AUs and other derived features. The activity of any AU is described by its active region (s) i.e. UL, ML, LL. An overview general observation of different actions and action locations of AUs for the corresponding actions is shown in Table.~\ref{Tab-1}. 


\begin{table}[]
\centering
\caption{Actions and action locations of AU}
\label{Tab-1}
\begin{tabular}{|c|c|c|c|}
\hline
\multirow{2}{*}{Action}                                                       & \multicolumn{3}{c|}{AU}      \\ \cline{2-4} 
                                                                              & UL         & ML        & LL  \\ \hline
\begin{tabular}[c]{@{}c@{}}Stand, Run,\\ Walk, jump,\\ turn back\end{tabular} & Head       & Hand      & Leg \\ \hline
Wave, Punch                                                                   & Head, Hand & Hand      & leg \\ \hline
Kick                                                                          & Head       & Hand, Leg & Leg \\ \hline
\end{tabular}
\end{table}


Table.~\ref{Tab-2} and Table.~\ref{Tab-3} give an instance of AUs, their acting regions and some other details of two actions walking and running respectively.

\begin{table}[h]
\centering
\caption{Action components and action level for Walking}
\label{Tab-2}
\begin{tabular}{|c|c|c|c|}
\hline
\multicolumn{4}{|c|}{{\bf Action: Walking}} \\ \hline
\multicolumn{1}{|l|}{{\bf Acting components}} & \multicolumn{1}{l|}{{\bf Upper level}} & \multicolumn{1}{l|}{{\bf Middle level}} & \multicolumn{1}{l|}{{\bf Lower level}} \\ \hline
Head & Yes & No & No \\ \hline
Hands & No & Yes & Yes \\ \hline
Legs & No & No & Yes \\ \hline
\multicolumn{4}{|l|}{\begin{tabular}[c]{@{}l@{}}Stride angle: (0-40) degree,\\ Location difference between two legs are very close to or less than (1/5)$\times$ H\end{tabular}} \\ \hline
\end{tabular}
\end{table}

\begin{table}[h]
\centering
\caption{Action components and action level for running}
\label{Tab-3}
\begin{tabular}{|c|c|c|c|}
\hline
\multicolumn{4}{|c|}{{\bf Action: Running}} \\ \hline
\multicolumn{1}{|l|}{{\bf Acting components}} & \multicolumn{1}{l|}{{\bf Upper level}} & \multicolumn{1}{l|}{{\bf Middle level}} & \multicolumn{1}{l|}{{\bf Lower level}} \\ \hline
Head & Yes & No & No \\ \hline
Hands & No & Yes & Yes \\ \hline
Legs & No & No & Yes \\ \hline
\multicolumn{4}{|l|}{\begin{tabular}[c]{@{}l@{}}Stride angle: (0-65) degree,\\ Location difference between two legs are very close to or less than (1/5)×H\end{tabular}} \\ \hline
\end{tabular}
\end{table}

It is understandable from above observation that the AUs and their action regions are main components to take the decision on actions. The action region is determined from the feature values of the corresponding AUs. Four AUs namely body, head, hands and legs are represented by BMC, head, (hand1, hand2) and (leg1 and leg2) respectively. Three other important derived components are SA, MD and BMC\_G. We have studied number of test case of different human actions collected from several publicly available human action datasets and observed that AUs are responsible to perform any action as we have seen values of AUs in different region for different actions. On the other hand the values of AUs are in a similar region when the actions are same however it is done by different actors. Thus we have develop a knowledge base for human action classification. The knowledge based includes the action regions of AUs and other derived features for different actions as shown in Table.~\ref{Tab-4}. 
Every AU has three properties viz. A) Moving values (MV): this value defines the amount of movement of the corresponding AU and this a temporal feature. The value of MV are any one out of {'0', and '1' and '2'}; '0' for no movement or static and '1' for slow movement and '2' for rapid movement. B) Moving direction(MDR): the direction of the corresponding AU. MDR can be of three different types {X, Y and XY} corresponding to horizontal, vertical and diagonal directions.  iii) Moving regions: The regions of AUs where action is taken place. Moving regions must have one of the three values viz. {UL,ML,LL). In Table.~\ref{Tab-4}, the symbol '--' used for don't care condition, that is the rule does not depend on that value, which is represented by '--'. The direction of any action is determine by the value of MD.

\begin{table*}[]
\centering
\caption{Knowledge base for action classicification}
\label{Tab-4}
\begin{tabular}{|l|l|l|l|l|l|l|l|}
\hline
Action    & (MV,MDR)  & \multicolumn{6}{l|}{Values of AU and other derived features (MV,MR)} \\ \hline
          & BMC\_G    & BMC      & Head     & Hand        & Leg        & SA       & MD       \\ \hline
Stand     & (0,--)    & (0,X)    & (0,X)    & (0,X)       & (0,X)      & --       & (-1,0,1) \\ \hline
Wave      & (0,--)    & (1,MID)  & (0,UP)   & (2,UP/MID)  & (0,LL)     & --       & (-1,0,1) \\ \hline
Punch     & (0,--)    & (1,MID)  & (1,UP)   & (2,UP/MID)  & (0,LL)     & (0-30)   & (-1,0,1) \\ \hline
Kick      & (0,--)    & (1,MID)  & (1,UP)   & (1,MID)     & (0,LL/MID  & (0-130)  & (-1,0,1) \\ \hline
Jump      & (2,Y/XY)  & (1,MID)  & (1,UP)   & (1,MID)     & (1,LL)     & --       & (-1,0,1) \\ \hline
Run       & (2,X/XY)  & (1,MID)  & (--,--)  & (--,--)     & (2,LL)     & (0-90)   & (-1,0,1) \\ \hline
Walk      & (2,X/XY)  & (1,MID)  & (--,--)  & (--,--)     & (2,LL)     & (0-60)   & (-1,0,1) \\ \hline
Turn back & (1, X/XY) & (1,MID)  & (1,UP)   & (1,MID)     & (1,LL)     & (0-30)   & (-1,0,1) \\ \hline
\end{tabular}
\end{table*}
 
\section{Results and analysis}
\subsection{Consideration:} The class of actions include the actions performed by a single human object with the body not upside down position . Thus the class of actions includes standing, walking, running, punching, kicking, jumping, jogging, clapping, waving etc. One important concern connected with HAR is that the estimation of the number of frames to determine human action. Most of the human actions are repetitive in nature i.e. the same cycle is continuing over a period of time. The time to complete a cycle is below a second in case of all actions considered in several datasets. We set the value of $k$ as the frame rate for this reason. The ultimate form of the STBPM feature is $28 \times 4$, which does not depend on $k$. In some of the cases more number of cycles may included in one test case, but that will not create any problem as STBPM includes features, which are extracted from all the sequences. Some of the actions are complicated and perform more than one actions  to complete a single one. For example, if there is any action of a running human who is taking rest by standing. This action is a combination of running and standing, but our proposed approach takes the most active action out of all performing actions. Thus for the above example, the proposed method will categorize the action as 'running'. The proposed method extracts features from silhouette of the objects collected from consecutive frames. So to validate the results, we need the help of those datasets having silhouette of the ground truth. The  performance evolution of the proposed action recognition framework  is done on two publicly available datasets: Weizmann ~\cite{35} and MuHAVi ~\cite{34}. All the datasets used here include silhouette sequences.
 
\subsection{Datasets:}
\subsubsection{Wizeman Dataset ~\cite{35}}
We conducted a series of experiments on the Weizmann Human Action Database available on-line ~\cite{35}. This is a very common dataset, many state-of-the-art approaches report performance on it thus allowing easy comparison. The database contains 90 low-resolution video and silhouette sequences ($180 \times 144$ pixels) that show 9 different people each performing 10 different actions, such as jumping, walking, running, skipping, etc. We have applied our algorithm except 'bend', which is violated the consideration upside down.
\subsubsection{MuHAVi ~\cite{34}}
To isolate the challenge of object detection, it is assumed that the segmentation problem (i.e. to obtain the silhouettes) has been solved. To address this, the dataset provides a sub-set of data that has been (painstakingly) manually annotated. The whole dataset can still be used in a combined segmentation/action recognition algorithm.
In MuHVAi-14 datasets, actions are classified into 14 different classes viz. "CollapseLeft", CollapsRight", "GuardToKick", "GuardToPunch", "KickRight", "PunchRight", "RunLeftToRight", "RunRightToLeft", "StandUpLeft", StandUpRight", "TurnBackLeft", "TurnBackRight", "WalkLeftToRight" and "WalkRightToLeft". We can classify the actions irrespective of direction and view points, but we cannot consider the "Collapse" and "StandUp" actions for our experiment. The extraction of AUs from silhouettes is an important area of our algorithm, but for those two actions when human body get squeezed in a smaller area, and AUs are not properly traceable by silhouette information. Other $10$ videos of the datasets are classified into $7$ actions in our considerations as the work is direction invariant. A brief of different action class and corresponding actions are shown in Table.~\ref{Tab-5}.

\begin{table}[]
\centering
\caption{Different action clss and actions of Wizeman and MuHVAi datasets}
\label{Tab-5}
\begin{tabular}{ll|l|l|}
\hline
\multicolumn{2}{|l|}{Wizeman Datases}                                                                                                                & \multicolumn{2}{l|}{MuHAVi Datasets}                                      \\ \hline
\multicolumn{1}{|l|}{\begin{tabular}[c]{@{}l@{}}Action \\ class\end{tabular}} & Action                                                               & \begin{tabular}[c]{@{}l@{}}Action\\  class\end{tabular} & Action          \\ \hline
\multicolumn{1}{|l|}{jack}                                                    & Jumping Jack                                                         & AC1                                                     & GuardToKick     \\ \hline
\multicolumn{1}{|l|}{jamp}                                                    & Jumping Forward                                                      & AC2                                                     & GuardToPunch    \\ \hline
\multicolumn{1}{|l|}{pjump}                                                   & Jumping in place                                                     & AC3                                                     & KickRight       \\ \hline
\multicolumn{1}{|l|}{run}                                                     & Running                                                              & AC4                                                     & PunchRight      \\ \hline
\multicolumn{1}{|l|}{side}                                                    & galloping sidewys                                                    & AC5                                                     & RunLeftToRight  \\ \hline
\multicolumn{1}{|l|}{skip}                                                    & \begin{tabular}[c]{@{}l@{}}skip one leg \\ while moving\end{tabular} & AC6                                                     & RunrightToLeft  \\ \hline
\multicolumn{1}{|l|}{walk}                                                    & walking                                                              & AC7                                                     & TurnBackRight   \\ \hline
\multicolumn{1}{|l|}{wave1}                                                   & waving an hand                                                       & AC8                                                     & TurnBackright   \\ \hline
\multicolumn{1}{|l|}{wave2}                                                   & waving two hands                                                     & AC9                                                     & WalkLeftToRight \\ \hline
                                                                              &                                                                      & AC10                                                    & WalkRightToLeft \\ \cline{3-4} 
\end{tabular}
\end{table}
\subsection{Evaluation}
The proposed methodology implemented a rule-base to classify different human actions. So the approach does not need any training for classification, but uses prior knowledge to develop the rule-base. We consider a few number of actions from a huge variety of human actions, but the effort is to construct the grammar, which can portray different human actions effectively. This technique is totally based on Human silhouette. So the proposed approach used the datasets where silhouette of corresponding video are provided. The following measures are used to measure the efficiency of the proposed method.
\subsubsection{Confusion matrix}
Confusion matrix gives a clear knowledge about actual and wrong classification of any classifier. Diagonal values of a confusion matrix determines the number true positive out of total number of action of that action class and the values other than diagonal values define the mis-classification. For example, as in table-6 in case of action class 'jump' '8' out of '9' action are successfully classified, and the remaining one is classified as action class 'run'. Confusion matrix for Wizeman and MuHVAi datasets are shown in Table.~\ref{Tab-6} and Table.~\ref{Tab-7} respectively. In case of Wizeman dataset we have confusions only for three action classes viz. 'jump', 'side' and 'skip'. All the three actions are confused with the action class 'run', which has a certain similarities with those three actions. On the other hand in case of MuHVAi datasets, the two action classes are confused with each other namely 'GuardToPunch' and 'GuardToKick'. The percentage of successful classification, which proves the efficiency of the proposed method are 95.06 and 93.75 for Wizeman and MuHVAi datasets respectively. This amount of success rate shows the efficiency of our technique.

\begin{table}[]
\centering
\caption{Confusion matrix for wizeman datasets}
\label{Tab-6}
\begin{tabular}{|l|l|l|l|l|l|l|l|l|l|}
\hline
	 &\rotatebox{90}{Walk} & \rotatebox{90}{Run} & \rotatebox{90}{Jump} & \rotatebox{90}{Pjump} & \rotatebox{90}{Side} &\rotatebox{90}{ Skip} &\rotatebox{90}{ Wave1} & \rotatebox{90}{Wave2} & \rotatebox{90}{Jack} \\ \hline
Walk   & 9/9     &     &      &       &      &      &       &       &      \\ \hline
Run    &      & 9/9    &      &       &      &      &       &       &      \\ \hline
Jump   &      &  1/9   &  8/9    &       &      &      &       &       &      \\ \hline
Pjump  &      &     &      &  9/9     &      &      &       &       &      \\ \hline
Side   &      & 1/9    &      &       &   8/9   &      &       &       &      \\ \hline
skip   &      &  2/9    &      &       &      &  7/9    &       &       &      \\ \hline
Wave1  &      &    &      &       &      &      &   9/9    &       &      \\ \hline
Wave2  &      &     &      &       &      &      &       &  9/9     &      \\ \hline
Jack   &      &     &      &       &      &      &       &       & 8/9     \\ \hline
\end{tabular}
\end{table}

\begin{table}[]
\centering
\caption{Cofusion matrix for MuHVAi datasets}
\label{Tab-7}
\begin{tabular}{|l|l|l|l|l|l|l|l|l|l|l|}
\hline
                & \rotatebox{90}{GuardToKick} & \rotatebox{90}{GuardToPunch} & \rotatebox{90}{KickRight} & \rotatebox{90}{PunchRight} & \rotatebox{90}{RunLeftToRight} & \rotatebox{90}{RunRightToLeft} & \rotatebox{90}{TurnBackLeft} & \rotatebox{90}{TurnBackRight} & \rotatebox{90}{WalkLeftToRight} & \rotatebox{90}{WalkRightToLeft} \\ \hline
GuardToKick     & 6/8            &   2/8           &           &            &                &                &              &               &                 &                 \\ \hline
GuardToPunch    & 2/8            &      6/8        &           &            &                &                &              &               &                 &                 \\ \hline
KickRight       &             &              &  8/8         &            &                &                &              &               &                 &                 \\ \hline
PunchRight      &             &              &           &      8/8      &                &                &              &               &                 &                 \\ \hline
RunLeftToRight  &             &              &           &            &    4/4            &                &              &               &                 &                 \\ \hline
RunRightToLeft  &             &              &           &            &                &               4/4 &              &               &                 &                 \\ \hline
TurnBackLeft    &             &              &           &            &                &                &    8/8          &               &                 &                 \\ \hline
TurnBackRight   &             &              &           &            &                &                &              &       8/8        &                &                 \\ \hline
WalkLeftToRight &             &              &           &            &                &                &              &               &        4/4         &                 \\ \hline
WalkRightToLeft &             &              &           &            &                &                &              &               &                 &          4/4       \\ \hline
\end{tabular}
\end{table}

\subsubsection{Misclassified frames (MCF)}
Our proposed approach is totally dependent of the correctness of AU localization. So the first thing we emphasize on accuracy of AU localization. The proposed approach gives good accuracy in terms of localization of the AUs for most of the frames of the video. We also apply the moving direction (MD) computation for each of the frames of different videos. MD is computed using two of the AUs viz. leg(s) and head and it is one of the main parameter to recognize the actions. The average mis-classification rate (AMR) of MDs in all videos are mentioned in  Table.~\ref{Tab-5} for the respective datasets and are shown in Table.~\ref{Tab-8}. For MuHAVi and Wizeman datasets, the MCF are 4.46\% and 3.20\% respectively. This mis-classification rate is very minimal with respect to the frame rate of the video as the decision is taken on the basis of majority response.

\begin{table}[]
\caption{Average miss-classification rate of MDs of the corresponding datasets}
\centering
\label{Tab-8}
\begin{tabular}{|l|l|l|}
\hline
Datasets & Actions & MAR \\ \hline
MuHAVi ~\cite{34} & 10 & 4.46 \\ \hline
Wiseman ~\cite{35} & 9 & 3.29 \\ \hline
\end{tabular}
\end{table}

\subsubsection{ Leave-one-actor-out (LOAO) and leave-one-sequence-out (LOSO) cross validation}
There are several human recognition approaches in related research domain, which are using LOAO and LOSO as the part of estimating the recognition ability of the different classification techniques. Any human action datasets contains several sequences of human actions performed by different actors. In case of LOAO, the classifier is trained by leaving the video of any one actor, that video will be used for testing purpose. On the other hand, in case of LOSO the classifier is trained by leaving any one sequence, which will be used for testing purpose. 

\par LOAO and LOSO can not be applicable for the proposed approach as the approach does not require training to classify human actions. On the other hand, the approaches use LOAO or LOSO for measuring their efficiency involving training and testing with a part of the dataset, which is not used for training, but in our proposed methods all the will be tested with out training. So the rate of successful classification of the proposed methodology can be easily comparable with the efficiency of the related research work. The results of applying our proposed method on Wizeman dataset our method with in comparison with that of the state of the art research works is shown in Table.~\ref{Tab-9} based on classification efficiency. 
The success rate of the proposed approach considering Wizeman dataset is $95.06\%$, which outperforms others. It can be inferred that though our technique doesn't yield hundred percent success rate, but the proposed work is a ready to be implemented one as it doesn't require any learning.

\subsubsection{ Identical training and testing actors, novel camera}
This is used for view invariance test of the algorithm for any dataset having video of the same actors from different angle. So in this case the training is done from the video taking through one camera, which is situated in a certain angle and testing the video which are taking through other camera, which is placed in the different angle. MuHVAi datasets provides multi-view data, which is exploited for novel camera test. Our proposed method does not depend on the camera view point when it is parallel to the object. We have tested all the videos, which are given in the dataset without training a single video, and the actions were recognized irrespective of the camera position. The results of our method and its comparison with the other state of the art is shown in Table.~\ref{Tab-10} for the MuHAVi dataset.  

\subsubsection{Identical training and test cameras, novel actors}
This testing phase is used for checking the robustness of the algorithm irrespective of actors i.e. the view angle of the actor in the test video differs with that of the actor in the tanning video.  MuHVAi dataset provides the videos from different angle of the same actor. We have applied our proposed method on all the videos irrespective of actors. To recognize actions of different actors, the proposed approach doesn't require training. Hence the proposed method qualifies the novel actor test. A comparative study of the results of our approach and that of the other related research works is shown in  
Table.~\ref{Tab-10} in context with classification efficiency on MuHAVi dataset for novel actor test.  
 \par The  accuracy rate of the proposed work over MuHAVi dataset is$93.75\%$, which is better than any other methods of the related research work. This performance also validate novel actor and novel camera test as we did not use any training mechanism.

\begin{table}[]
\caption{Compartive study of several methodologies of the related research with the proposed approach for wizeman datasets}
\centering
\label{Tab-9}
\begin{tabular}{|l|l|l|l|l|}
\hline
Approach & Input & Actions & Evaluations & Rate \\ \hline
Ikizler and Duyugulu ~\cite{10} & Silhouette & 9 & LOSO & 100 \\ \hline
Tran and Sorokin ~\cite{11} & Silhouette & 10 & LOSO & 100 \\ \hline
Eweiwi et al. ~\cite{12} & Silhouette & 10 & LOSO & 100 \\ \hline
Harnandez et al. ~\cite{13} & Images & 10 & LASO & 90.3 \\ \hline
Cheema et al.  ~\cite{6} & Silhouette & 9 & LOSO & 91.6 \\ \hline
Chaaraoui et al. ~\cite{8} & Silhouette & 9 & LOSO & 92.8 \\ \hline
Proposed & Silhouette & 9 & No Training & 95.06 \\ \hline
\end{tabular}
\end{table}

\begin{table}[]
\caption{Compartive study of several methodologies of the related research with the proposed approach for MuHVAi datasets}
\centering
\label{Tab-10}
\begin{tabular}{|l|l|l|l|l|}
\hline
Approach & Input & Actions & Evaluations & Success Rate \\ \hline
Singh et al. ~\cite{7} & Silhouette & 14 & LOSO & 82.4 \\ \hline
Eweiwi et al. ~\cite{12} & Silhouette & 14 & LOSO & 91.9 \\ \hline
Cheema et al. ~\cite{6} & Silhouette & 14 & LOSO & 86.0 \\ \hline
Chaaraoui et al. ~\cite{8} & Silhouette & 14 & LOSO & 92.8 \\ \hline
Proposed & Silhouette & 10 & No Training & 93.75 \\ \hline
\end{tabular}
\end{table}

\begin{table}[]
\caption{ results on MuhVAi datasets for novel actor test}
\centering
\label{Tab-11}
\begin{tabular}{|l|l|l|l|l|}
\hline
Approach & Input & Actions & Evaluations & Success Rate \\ \hline
Singh et al. ~\cite{7} & Silhouette & 14 & LOSO & 61.8 \\ \hline
Eweiwi et al. ~\cite{12} & Silhouette & 14 & LOSO & 77.9 \\ \hline
Cheema et al. ~\cite{6} & Silhouette & 14 & LOSO & 73.5 \\ \hline
Chaaraoui et al. ~\cite{8} & Silhouette & 14 & LOSO & 82.4 \\ \hline
Proposed & Silhouette & 10 & No Training & 93.75 \\ \hline
\end{tabular}
\end{table}

\section{Conclusion}

The present work proposes a new spatio-temporal feature extraction technique coined as STBPM and a rule based logic termed as RAC to classify human actions. Feature extraction is done on silhouette of the foreground for $k$ number of consecutive frames. Our proposed technique successfully localizes AUs and determines human actions with high accuracy except the action with head upside down,  sitting and lying. Our technique doesn't required any training and is also independent of the camera view angle. Experimental results involving publicly available datasets shows that our technique outperforms the other state of the art techniques in terms of success rate of HAR. In future, we wish to extend our work for detecting action in the sitting and lying conditions making the detection process more extensive.  

\section{Acknowledgment} 
This work is funded by DST, Ministry of Science and Technology, Government of India through INSPIRE project.

\end{document}